\documentclass[runningheads]{llncs}

\usepackage{amsmath,amssymb,amsfonts}
\usepackage{graphicx}
\usepackage{subcaption}
\usepackage{textcomp}
\usepackage{url}
\usepackage{amsmath}
\usepackage{longtable}
\usepackage{xcolor}
\usepackage{multirow}
\usepackage{algorithm}
\usepackage{algpseudocode}
\def\BibTeX{{\rm B\kern-.05em{\sc i\kern-.025em b}\kern-.08em
    T\kern-.1667em\lower.7ex\hbox{E}\kern-.125emX}}

\usepackage{listings}

\usepackage[style=ieee,maxnames=2,minnames=1,backend=biber,bibstyle=ieee,citestyle=numeric-comp]{biblatex}
\newcommand{\round}[1]{\ensuremath{\lfloor#1\rceil}}

\begin{document}
\title{AI in Space for Scientific Missions: Strategies for Minimizing Neural-Network Model Upload\thanks{This work is supported by the European Commission, with Automatics in Space Exploration (ASAP), project no. 101082633.}
}
\titlerunning{AI in Space for Scientific Missions}
\authorrunning{J. Ekelund et al.}
\author{Jonah Ekelund$^1$, Ricardo Vinuesa$^1$, Yuri Khotyaintsev$^2$, Pierre Henri$^3$, Gian Luca Delzanno$^4$, Stefano Markidis$^1$}
\institute{
\textit{$^1$ KTH Royal Institute of Technology, Stockholm, Sweden}\\
\textit{$^2$ Swedish Institute of Space Physics, Uppsala, Sweden}\\
\textit{$^3$ CNRS Researcher, Lagrange, OCA, Nice \& LPC2E, Orléans, France}\\
\textit{$^4$ Los Alamos National Laboratory, Los Alamos, NM, USA}\\
}
\maketitle
\begin{abstract}
  Artificial Intelligence (AI) has the potential to revolutionize space exploration by delegating several spacecraft decisions to an onboard AI instead of relying on ground control and predefined procedures. It is likely that there will be an AI/ML Processing Unit onboard the spacecraft running an inference engine. The neural-network will have pre-installed parameters that can be updated onboard by uploading, by telecommands, parameters obtained by training on the ground. However, satellite uplinks have limited bandwidth and transmissions can be costly. Furthermore, a mission operating with a suboptimal neural network will miss out on valuable scientific data. Smaller networks can thereby decrease the uplink cost, while increasing the value of the scientific data that is downloaded. In this work, we evaluate and discuss the use of reduced-precision and bare-minimum neural networks to reduce the time for upload. As an example of an AI use case, we focus on the NASA's Magnetosperic MultiScale (MMS) mission. We show how an AI onboard could be used in the Earth's magnetosphere to classify data to selectively downlink higher value data or to recognize a region-of-interest to trigger a burst-mode, collecting data at a high-rate. Using a simple filtering scheme and algorithm, we show how the start and end of a region-of-interest can be detected in on a stream of classifications. To provide the classifications, we use an established Convolutional Neural Network (CNN) trained to an accuracy >94\%. We also show how the network can be reduced to a single linear layer and trained to the same accuracy as the established CNN. Thereby, reducing the overall size of the model by up to 98.9\%. We further show how each network can be reduced by up to 75\% of its original size, by using lower-precision formats to represent the network parameters, with a change in accuracy of less than 0.6 percentage points.
\keywords{ Space Exploration, Artificial Intelligence in Space, Compressed Neural Networks, Neural Network Parameter Upload}
\end{abstract}

\section{Introduction}\label{introduction}
Artificial-intelligence (AI) and Machine-learning (ML) methodologies and tools are changing and will change space exploration by offloading several spacecraft decisions to an onboard AI instead of solely relying on ground control and predefined procedures~\cite{kothariFinalFrontierDeep2020a}.

An early example of the use of neural-networks in space is the CNES DEMETER (Detection of Electro-Magnetic Emissions Transmitted from Earthquake Regions) satellite (2004-2010), devoted to the investigation of the ionospheric disturbances due to seismic and volcanic activities.
DEMETER implemented a TDNN (Time Delay Neural Network) in the onboard software for the automatic and systematic detection and characterization of all whistler-like signals~\cite{ElieDEMETER1999} encountered by DEMETER along its orbit~\cite{ParrotDEMETER2019}. To our knowledge, this is the first on-board neural network dedicated to the scientific processing of wave data.

A spacecraft equipped with an AI processing unit could detect which part of space is traversing. Using this information, the measurement sampling rate could be increased in areas of strong scientific interest while, conversely, neglecting data that might not be of interest. In the case of space exploration of the Earth’s magnetosphere, for instance, a so-called \textit{burst mode}, an increased sampling rate of on-board instruments~\cite{pfaffOverviewFastAuroral2001,lakeyOptimisationSolarOrbiter2016,bakerMagnetosphericMultiscaleInstrument2016}, can be triggered by in-situ data analysis using an on-board AI engine. For instance, an onboard AI module can identify transition regions in the Earth’s magnetosphere, such as the magnetopause, where day-side magnetic reconnection might occur, the bow-shock region and turbulence generated by ion-foreshock and in general, anomaly detection to measure rare events. All these events might interest space scientists on the ground. Automatic detection will increase the value of the data received by the scientists, since high-interest areas will be prioritized for collection and downlink.

In the first implementation, the onboard spacecraft AI will likely execute in a separate computing unit, radiation-hardened or replicated for redundancy and fault tolerance, here called ML/AI Processing Unit (MAP), running an inference engine (a neural network model with only prediction capabilities). The MAP will contain an accelerator to accelerate tensorial or vector operations, which are key calculations for inference: Nvidia tensor cores~\cite{markidis2018nvidia}, Google Tensor Processing Units~\cite{jouppi2017datacenter} and Intel Movidius Myriad Vision Processing Units~\cite{rivas2018exploring} are a few commercial examples. The MAP comes with pre-installed neural-network architecture, weights and biases (the so-called model) in a standardized format, such as Open Neural Network Exchange (ONNX)~\cite{onnx2024}. During the space mission, the availability of new space observations from the instruments and re-calibration of onboard instruments will allow new extensive training of the neural networks on the ground and preparation of new neural network architecture, weights and biases. The new model can be uploaded from the ground by uploading weights and biases obtained by training on the ground. However, the upload time largely depends on the amount of data transmitted to the spacecraft. In fact, the bandwidth upload is limited and overall, communication with the spacecraft is costly in terms of resources and time. As will be shown in this work, the neural network parameter upload time might vary from minutes to hours and it is therefore critical to minimize the amount of data that needs to be transmitted when updating the model.

The main challenge for neural-network model upload is that satellite uplinks have limited bandwidth and communication. Furthermore, a mission operating with a sub-optimal neural network will not be accurate, potentially missing valuable scientific data, or wasting resources on uninteresting data. Smaller networks, in terms of architecture and parameter precision, can thereby decrease the uplink cost while increasing the value of the downloaded scientific data. There are many existing techniques for reducing the size of a network, for example pruning and quantization.

When pruning a network, a trained model is examined for neurons and parameters which are redundant or have little effect on the final output~\cite{LIANG2021370}. These can be removed to decrease the size of the network in the case of neurons or allow for more compact storage formats to be used if parameters are unnecessary and set to zero.

Another approach to reduce the size of a neural-network is quantization. For neural-networks, quantization involves using lower-precision representations for the parameters which, in addition to reducing the storage size, have been shown to increase the inference speed. Commonly, the networks are quantized from 32-bit float to 16-bit or 8-bit representations~\cite{LIANG2021370}; however, even 1-bit networks are possible~\cite{gunturk2023approximation}. Depending on the level of quantization, this also has the possibility to reduce the power usage, both through the use of less power intensive operations, but also by requiring fewer memory accesses~\cite{hubaraQuantizedNeuralNetworks2018}.

In this work, we evaluate and discuss two main approaches for neural-network size reduction in the context of reducing the amount of data that needs to be transferred to the spacecraft. First, the network size can be reduced by using fewer and smaller layers. Second, quantizing the network's parameters from the standard 32-bit to 16-bit, standard IEEE format or new emerging brain floating point (BFloat16), or basic 8-bit format.

The goal of this work is to evaluate strategies to minimize the communication cost while retaining accuracy and performance of the on-board neural-network engine. These strategies include the usage of bare-minimum neural architectures with just a few layers and neural units and the usage of low-precision format for encoding the neural network weights and biases. To assess the communication costs and neural network performance, we take the current NASA Magnetosperic MultiScale (MMS) mission, exploring the multi-scale nature of magnetic reconnection at electron scales in the Earth magnetosphere, as an example of mission for which AI-on board can enable burst modes automatically and increase instrument measurement rates~\cite{bakerMagnetosphericMultiscaleInstrument2016}. As a use case, we employ the automatic recognition of a region of interest in the day-side Earth magnetosphere using data from one of the on-board instruments, called FPI (Fast Plasma Investigation), for the observation of ion distributions at different energy levels~\cite{pollockFastPlasmaInvestigation2016}. Different models are trained, then file sizes and inference accuracy are evaluated for different neural network architectures and precisions. The main contributions of this work are the following:
\begin{enumerate}
    \item We present two practical uses of machine learning and neural networks onboard spacecraft, prioritization of data for downlinking and identifying region of interests.
    \item We present a simple way of filtering classification output from a region classifier (in this work, neural-networks) and using the classification to detect a region of interest.
    \item We show two reduced networks with similar performance to an existing network on the same dateset, decreasing the overall size and thereby the time required to upload the models to a satellite by up to 98.9\%.
    \item We show how the precision of the network parameters can be reduced, to 16- and 8-bit formats, with minimal change in the accuracy of the network predictions; Further decreasing the model size by up to 75\%.
\end{enumerate}

\section{Background}\label{background}
In this work, we use the MMS mission as a use-case for our study. MMS was launched on March 15th, 2015, with the mission of investigating magnetic reconnection in the Earth's magnetosphere boundary regions and enlightening the processes at electron and ion scales. MMS consists of four spacecrafts flying in tetrahedral formation: this allows to obtain the gradient of the various plasma and field measurements~\cite{bakerMagnetosphericMultiscaleInstrument2016}. There are two phases of the MMS mission. In Phase 1, the orbits are designed to maximize the time spent in the magnetopause on the Earth's day side. For Phase 2, the orbits are designed to fly the spacecraft through the magnetotail on the Earths night side~\cite{fuselierMagnetosphericMultiscaleScience2016}. In this work, for sake of simplicity, we focus on Phase 1, as well as on data and techniques designed to automatically detect day-side Earth Magnetosphere.

The scientific instruments onboard spacecraft today are capable of generating large amounts of data, which often exceeds the amount that can be efficiently downloaded. In the MMS mission, only $\sim$2\% of the high-rate data can be downloaded while using 75\% of the available downlink bandwidth~\cite{bakerMagnetosphericMultiscaleInstrument2016}. Furthermore, due to the location of ground stations, orbital dynamics and onboard resource limitations such as thermal generation and power consumption, the communication with the spacecraft is often further limited to different communication windows. In MMS Phase 1 these were between 15 and 80 minutes, in different parts of the orbits~\cite{fuselierMagnetosphericMultiscaleScience2016}.

Note that, deep-space missions, like ESA's Solar Orbiter, have the added difficulty that the downlink rate changed significantly depending on the distance of the spacecraft from Earth. This had to be considered when designing the mission orbit for the Solar Orbiter, minimizing the risk of overflowing the onboard mass memory and loosing valuable scientific data~\cite{lakeyOptimisationSolarOrbiter2016}.

A common approach to deal with this limitation, is to only collect low-rate survey data for most of the orbit and limit the collection of high-rate (often called "burst") data to specific regions of interest (ROI). These bursts of high-rate data collections can be triggered by onboard algorithms or events, like in the case of the FAST satellite and the Solar Orbiter~\cite{pfaffOverviewFastAuroral2001,lakeyOptimisationSolarOrbiter2016}. Alternatively, like for the MMS mission, the ROI can be specified as a longer part of the orbit, then the collected data can be sorted into a prioritized download queue based on the likelihood of containing interesting events. For MMS, this is a complicated process involving downloading onboard calculated quality indicator values from all four spacecraft, to then processing these on ground to produce the download prioritization which is then uploaded to the spacecraft during a following communication window. In addition to this, survey data is downloaded for a scientist to evaluate and change the prioritization~\cite{bakerMagnetosphericMultiscaleInstrument2016}.

As scientific missions are often more concerned with optimizing the amount of data that can be downlinked from the spacecraft, a limitation which is often not discussed in literature is the uplink. In the case of MMS the uplink data rate is only 2 kbps~\cite{raphaelCommandAmpData2014} and with communications windows between 15 and 80 minutes only 225 kB to 1.2 MB can be uploaded depending on the window; Larger files will have to be split between multiple windows. This does not account for any protocol headers that need to be added. In general, as frequency bandwidth is a limited resource, it needs to be utilized efficiently~\cite{ituconst2023} and the spacecraft uplink rates are often significantly lower than the downlink rates~\cite{menzelDesignVerificationPerformance2023,kobayashiIrisDeepSpaceTransponder}.

For traditional space missions this is not an issue since mainly commands are sent to the spacecraft. However, scientific missions are designed to study new phenomena where there is a lack of existing data usable for training neural networks. Therefore, missions which utilize pretrained AI models will have to update the models when real data becomes available. As we will see in this work, the upload time can be substantial even for small networks, when the uplink data rate is limited.

\subsection{Onboard ML/AI Acceleration}
\begin{figure}[t]
    \centering
    \includegraphics[width=0.8\linewidth]{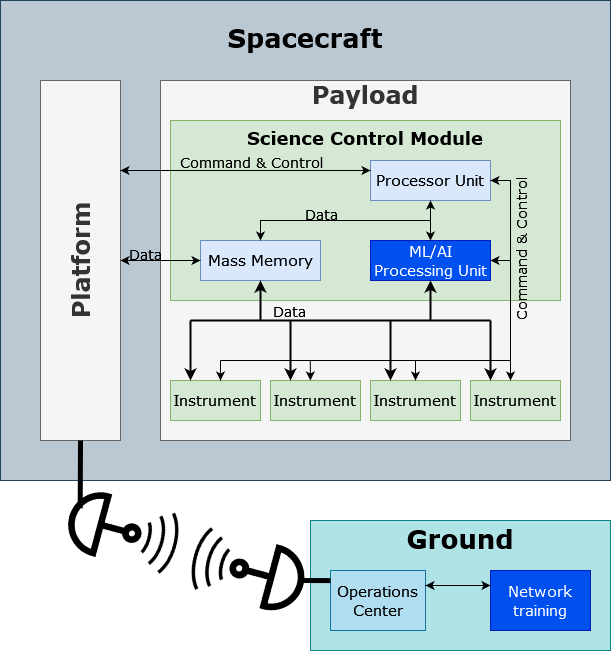}
    \caption{Schematic view of an ML/AI acceleration onboard a spacecraft for scientific-missions.}
    \label{fig:onboard_ai_acceleration}
\end{figure}

Current scientific space missions do not include AI-based modules for on-board data analysis for automatic or assisted decision. However, we envision the use of an overall architecture, similar to the one shown in Fig.~\ref{fig:onboard_ai_acceleration}, for a multi-instrument scientific mission. This architecture would adopt a separation of concern in the design, dividing the science payload, the management of the scientific instrumentation, from the spacecraft platform in charge of the management/control of the spacecraft. The Science Control Module (SCM) contains the processor unit for controlling the science payload, the Mass Memory Unit (MMU) for storing the science data and the ML/AI Processing Unit (MAP) for running the Inference Engine and supporting software. Depending on the use case, the Instruments will either store data to the MMU for the MAP to read and process, or stream data directly to the MAP, for a live processing pipeline. The MAP will create a number of key-values which can be used for prioritizing data or triggering burst modes. Depending on the level of instrument autonomy, the key-values can either be broadcast to anyone listening or read by the Processor Unit for use in controlling the instruments.

The implementation of the SCM can be either as a single board or as multiple boards with a backplane (motherboard) connecting them. This second variant could be used for a more modular design where multiple MMUs or MAPs can be added or removed depending on the mission needs.

In the first implementations the MAP will only run an Inference Engine, while the training will be performed on-ground and new models will be uploaded to the spacecraft when needed as shown in Fig.~\ref{fig:onboard_ai_acceleration}.

\subsection{AI-Based Methods \& Reduced Precision}\label{convolutional-neural-networks}
In this work, we focus on the usage of Convolutional Neural Networks (CNN), which are neural-networks often used for image analysis and have been proven to provide high accuracy in classifying different plasma regions in Earth's magnetosphere~\cite{olshevskyAutomatedClassificationPlasma2021}. They have one or more convolutional layers at the beginning of the network, which are used to extract key features from the images before they are fed into the network layers. This is then followed by one or more network layers.

For the communication to the spacecraft and upload, it is likely that a standardized data, such as ONNX, will be used. ONNX is an open-source format for representing AI and ML models. It is designed to provide a common model representation to facilitate exchange between different AI/ML frameworks. This enables developers and researchers to choose the best tool for each stage of a project~\cite{onnx2024}.

\begin{figure}[t]
    \centering
    \includegraphics[width=\linewidth]{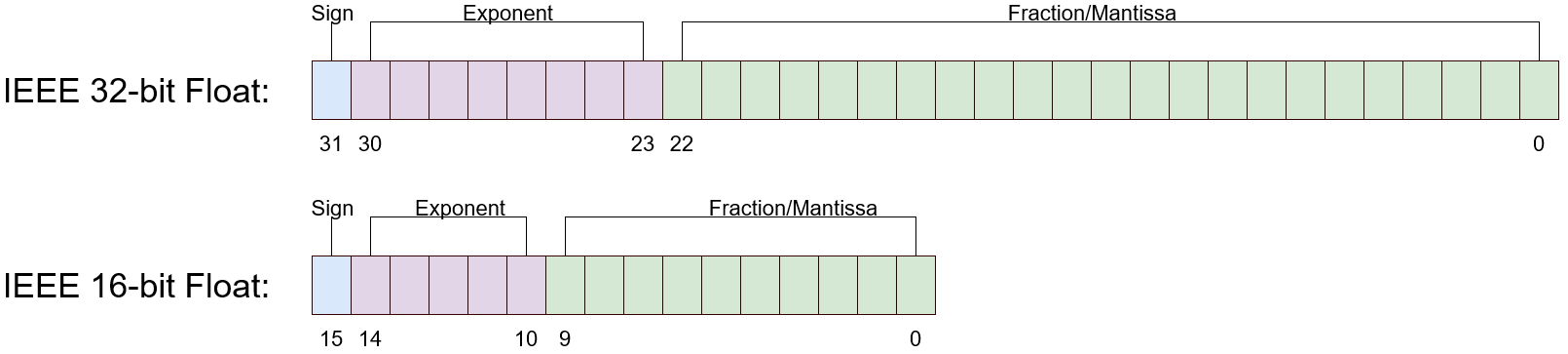}
    \caption{IEEE 32- and 16-bit floating-point formats.}
    \label{fig:float_formats}
\end{figure}

An important way to reduce the amount of data to be uploaded is to use reduced-precision formats. A common way of representing numbers in computers today is floating-point, which takes any real number splits it into a sign, exponent and fractional part, as shown in Fig~\ref{fig:float_formats}. This way of representing the number allows for storage of both very large numbers, such as the mass of stars and very small numbers, like the charge of an electron. A common standard for floating-point is the IEEE~\cite{alma99482816002456} floating-point standard, the 32-bit and 16-bit precisions are depicted in Fig.~\ref{fig:float_formats}. An alternative to the IEEE 16-bit floating-point representation is BFloat16. BFloat16 uses the same number of bits for the exponent as IEEE 32-bit, but reduces the fractional part to 7-bits. Note that this keeps the dynamic range of the IEEE 32-bit numbers~\cite{bfloat_web} and has been shown to be beneficial when training neural-networks using 16-bit number representations~\cite{kalamkarStudyBFLOAT16Deep2019}.

\section{Methodology}\label{use-cases-for-scientific-space-missions.}

Based on the limitations in downlink capabilities of scientific missions, we can identify two use-cases where a CNN similar to the one presented in Ref.~\cite{olshevskyAutomatedClassificationPlasma2021} can be used: \\

\noindent   \textbf{1. Selective downlink of scientific data onboard the spacecraft.} To selectively downlink data with high-value first, the classification output from the CNN could be used as an input to a prioritization algorithm, similar to how burst-data segments are selected in the MMS mission. The algorithm could put all the data in a prioritized download queue, where the lower-priority data could also be overwritten if the memory is full. The simplest example of this would be to prioritize all the data from one region. However, the algorithm could also be more complex, considering other data quality indicators or orbit information. For an Earth-orbiting mission, such as MMS, unless you want near real-time data, it might not make sense to put the CNN on the spacecraft itself because the time between two consecutive ground contacts is relatively short. Instead, for these missions the CNN could be used in on survey data downloaded to ground at a lower time resolution, similar to the \textit{Scientist in the Loop} (SITL) strategy used in MMS. However, for missions outside near-Earth orbits, such as missions to Mars or Jupiter, the communication delay can be large and the orbital dynamics can cause long periods without ground contact. Therefore, it may be preferable that the spacecraft starts transmitting the most relevant data as soon as ground contact has been established and not download survey data for evaluation on the ground. \\

\noindent  \textbf{2. Identifying region of interest (ROI) for high-rate and high-precision data collection.} Identifying the ROI to trigger a burst mode could be done for different reasons. First, similar to the data prioritization, identification of ROI could be done to limit the data collection due to storage and downlink limitations, only collecting data of high value. ROI could also be used to limit the use of expendable resources required for the data collection, such as the Indium in MMS ASPOC instrument~\cite{torkarActiveSpacecraftPotential2016}. Both involve detecting the boundary between two plasma regions. % basic strategies for boundary detection are elaborated upon in Sec. \ref{sec:boundry_detection}.

\subsection{Classification and ROI detection}\label{sec:boundry_detection}
For the first use-case (data download prioritization) the classification output from the CNN could be used as a way to determine how much of the data in a given data segment is from each region. Then data containing a larger percentage of data from a plasma region of interest can then be prioritized higher. The classification could also be used in more advanced segmentation algorithms to create segments of data containing only one classification.

The CNN networks presented in Section~\ref{sec:network} are evaluating each instance of the input data by itself and do not consider the previous input. However, when a network is running onboard a spacecraft with a continuous stream of data from the instrument, there is a dependency on the previous classification. For example, the spacecraft will not pass directly from the magnetosphere into the solar wind. We can add this dependency on previous data to the classification output by applying an exponentially weighed moving average to the classification output
\begin{equation}
    y_{i} = (1-\alpha)y_{i-1} + \alpha \cdot C_i
\end{equation}
then to obtain the classification we round to the nearest integer: $C_{i} = \round{y_{i}}$.

\begin{algorithm}[t]
\caption{Algorithm for detecting a region of interest.}\label{alg:roi_classfication}
\begin{algorithmic}[1]
\State $threshold \gets THRESHOLD$
\State $decay \gets DECAY$
\State $regions \gets REGIONS$
\State $indicator \gets 0$
\State $in\_roi \gets \text{\textbf{False}}$
\Function{check\_roi}{$c$}
    \If{$in\_roi$}
        \If{$c$ in $regions$}
            \State $indicator \gets indicator \cdot (1-decay)^{-1}$
            \If{$indicator \geq 1.0$}
                \State $indicator \gets 1.0$
            \EndIf
        \Else
            \State $indicator \gets indicator \cdot (1-decay)$
            \If{$indicator \le threshold$}
                \State $indicator \gets 0.0$
                \State $in\_roi \gets \text{\textbf{False}}$
            \EndIf
        \EndIf
    \Else
    \If{$c$ in $regions$}
            \State $indicator \gets 1.0$
            \State $in\_roi \gets \text{\textbf{True}}$
        \EndIf
    \EndIf
    \State \Return $in\_roi$
\EndFunction
\end{algorithmic}
\end{algorithm}

With the filtered classification, we can use a simple algorithm, Algorithm~\ref{alg:roi_classfication}, to detect the start and end of a region of interest. The algorithm detects if the given classification (\texttt{c}) is one of the regions of interest and returns \texttt{True} if it is. This is a quick response to ensure that we mark a ROI quickly for any associated actions to be performed. It relies on the previous filtering to remove any spurious classifications.

The algorithm has an indicator value that is set to one when entering a region of interest. It will then decay according to a configured value if a classification is not in our region of interest and when the indicator reaches a configured threshold value the classification will be marked as not in ROI. If the indicator has not decreased below the threshold and a classification is in the region of interest, the indicator will increase with the same decay term. This part of the algorithm ensures that transition regions with rapidly changing classifications are still captured.

\subsection{Data \& Pre-Processing}\label{data}
\begin{figure}[ht]
    \centering
    \includegraphics[width = \linewidth]{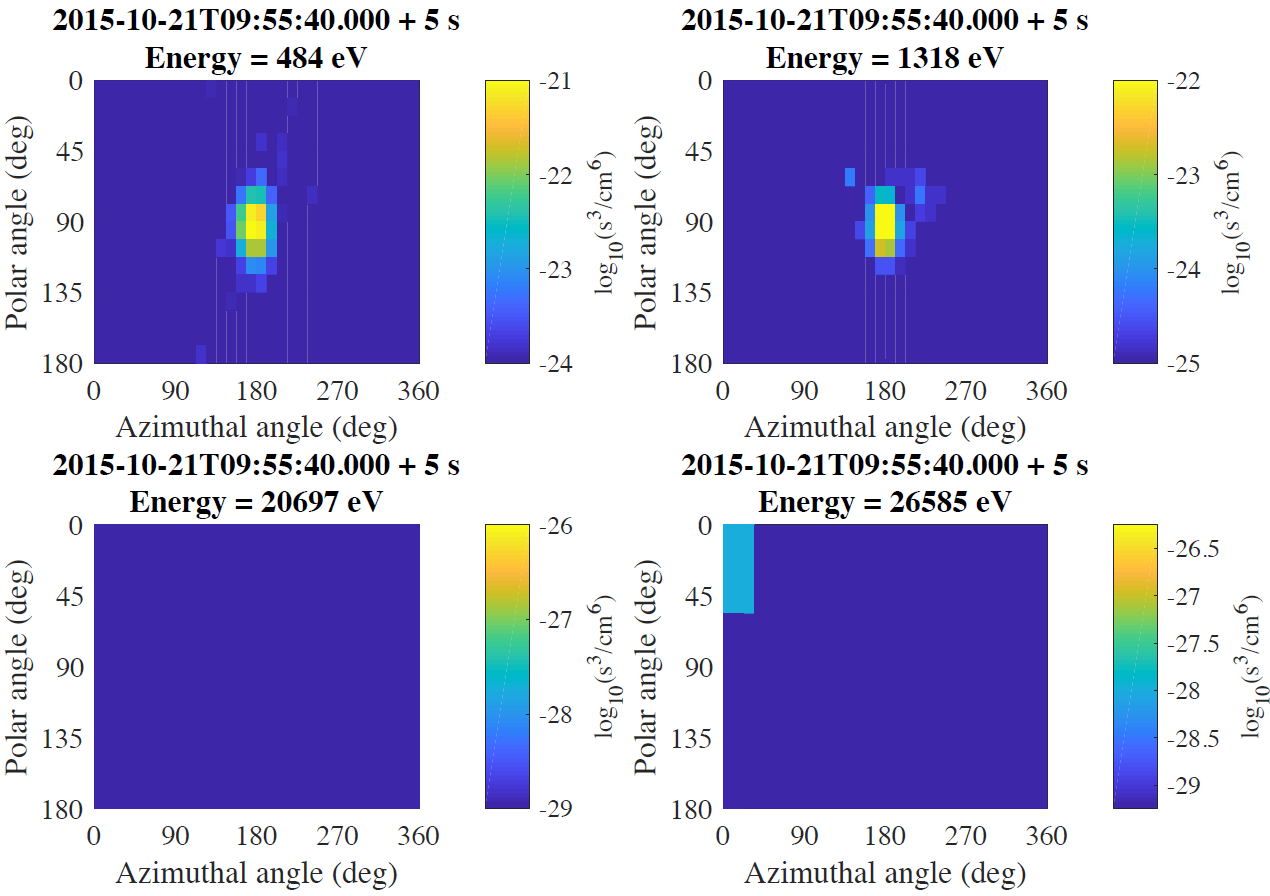}
    \caption{Example of skymaps for four energy levels from the Solar Wind region, data from MMS1. In the top most plots, we can see how the particle distribution for the Solar Wind appears as a Maxwell-Boltzmann distribution in the middle of the plot due to the rotation of the data along the $\phi$-axis.}
    \label{fig:ion_disp_sw}
\end{figure}

The Data for this work is ion energy distribution data from the MMS Fast Plasma Investigation instrument (FPI)~\cite{pollockFastPlasmaInvestigation2016} and was obtained from~\textcite{MMCScienceData2024}. The ion distribution data from FPI is collected in 32 energy levels with coverage from 10eV to 30keV. Spatially the data is collected in 16 polar angle bins and 32 azimuthal bins giving a sky map of the ion distribution around the spacecraft~\cite{pollockFastPlasmaInvestigation2016}. Fig~\ref{fig:ion_disp_sw} show a subset of the available energy-levels for Solar Wind data.

\textcite{olshevskyAutomatedClassificationPlasma2021} labeled the MMS data from November 2017 and December 2017 as one of four categories:
\begin{tabular}{rl}
-1. & Undefined/Unknown \\
0. & Solar Wind (SW) \\
1. & Ion foreshock (IF) \\
2. & Magnetosheath (MSH) \\
3. & Magnetosphere (MSP)
\end{tabular}

Using these labels, together with the corresponding MMS data, we generated two datasets, one for MMS data collected in November 2017 and one for December 2017. For each set, 10,000 samples were randomly selected from each of the four categories, SW, IF, MSH and MSP, generating a total 40,000 samples for each dataset. Samples labelled as Unknown were not included in the datasets. The November data set, was used as a training-set and the December set was used as a test-set.

The pre-processing of the ion distribution data was done in four steps:
\begin{enumerate}
    \item Limit the values in each sample to a given range
    \item Take log10 of all values in each sample
    \item Normalize each sample
    \item Perform a roll along $\phi$-axis in the data matrix
\end{enumerate}
which is slightly different from what was done in Ref.~\cite{olshevskyAutomatedClassificationPlasma2021}.

In the first step, any value in the data outside a range defined by a low and high threshold value was replaced with a corresponding threshold value. The range was chosen to be $[10^{-28}, 10^{-17}] \text{ (s}^3/\text{cm}^6$) based on the maximum and minimum values of the input data in the training set. This is done for two reasons, firstly to not have any zero values for the logarithm operation in the next step. Secondly, to ensure that all values will be between 0 and 1 after the normalization step. The logarithm step is the same as in~\cite{olshevskyAutomatedClassificationPlasma2021}.
In the third step, we normalize each sample between the logarithm of the threshold values. In practice, this will mean subtracting \(-28\) and dividing with \(11\) (the difference between the logarithm of the high and low threshold values.) The fourth and last step are the same as in Ref.~\cite{olshevskyAutomatedClassificationPlasma2021} and is done to put the solar wind beam in the center of the measurement box, as can be seen in Fig.~\ref{fig:ion_disp_sw}.

\subsection{CNN Architectures} \label{sec:network}
We use a CNN with the network topology presented by \textcite{olshevskyAutomatedClassificationPlasma2021} as the basis in this work. It classifies the different regions in the Earth's plasma environment into one of four categories: SW, IF, MSH or MSP.

\begin{figure}[t]
    \centering
    \includegraphics[width=0.8\linewidth]{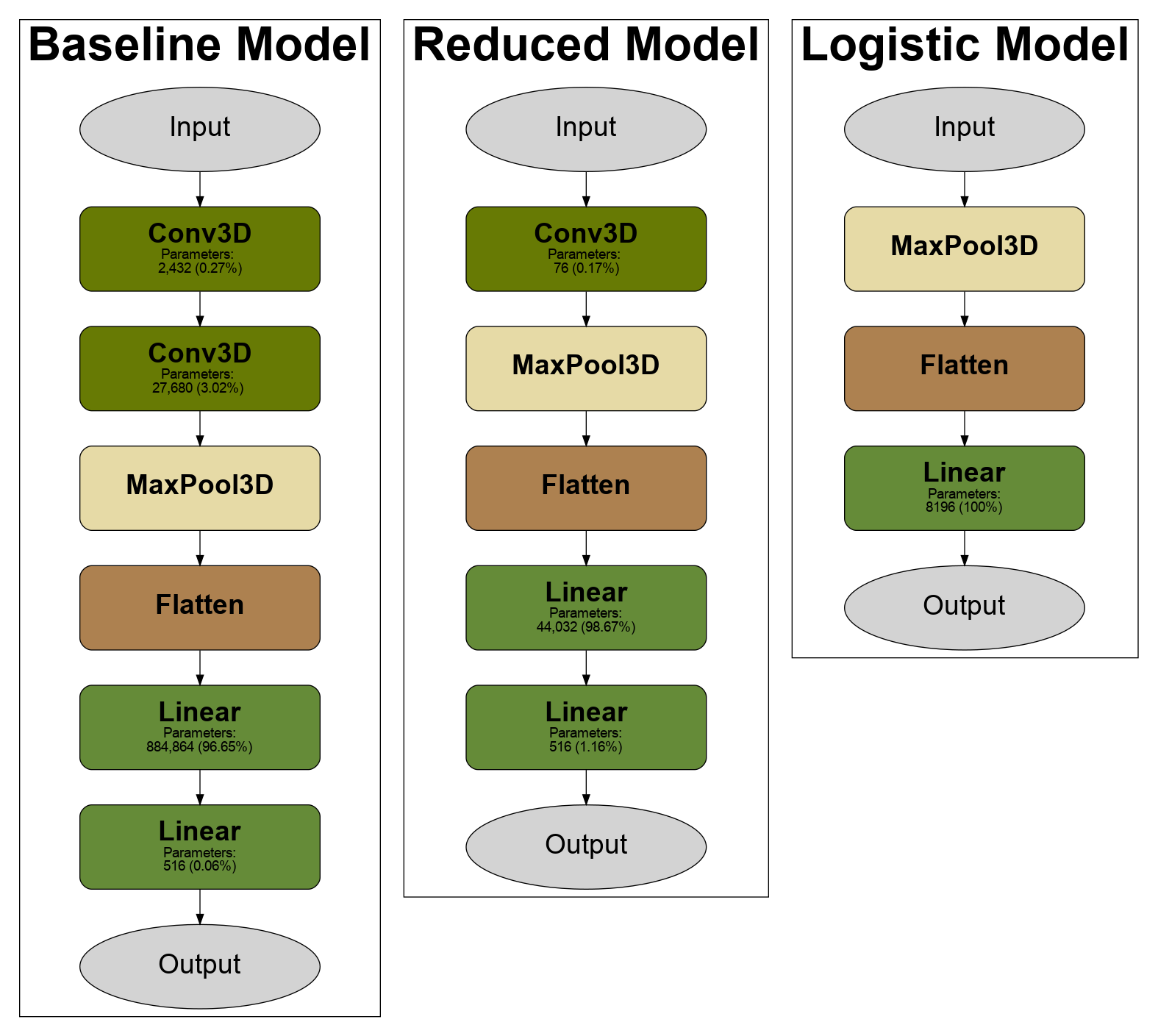}
    \caption{CNN architectures evaluated in this study.}
    \label{fig:cnn_network}
\end{figure}

This is a small CNN, see Fig.~\ref{fig:cnn_network}, consisting of two convolution layers and a max pool layer which reduces the initial 16,384 inputs to 6912. This is followed by two fully connected (linear) layers, the first with a ReLU activation function and the second with a soft max activation. A full summary of the network configuration is presented in Tab.~\ref{tbl:cnn_network_config}. The output from the network is the likelihood that the input data is from one of the four plasma regions. We implement this network in PyTorch and train it to identify the four different regions.

\begin{table}[]
    \centering
    \caption{Layer configuration for the CNNs}
    \label{tbl:cnn_network_config}
    \begin{tabular}{c|c|c|c}
         Layer &  Configuration & Baseline CNN & Reduced CNN \\
         (Type) &  &  & \\\hline
               & In Channels & 1 &  1 \\
           1   & Out Channels & 32 & 1\\
        (Conv3D)   & Kernel size & (5,3,5) & (5,3,5) \\
               & Stride & (2,1,2) & (2,1,2)\\
               & Padding & 0 & 0\\ \hline
               & In Channels & 32 &   \\
           2   & Out Channels & 32 & \\
        (Conv3D)   & Kernel size & (3,3,3) & Layer not used \\
               & Stride & (1,1,1) & \\
               & Padding & 0 & \\ \hline
            3  & Kernel size & 2 & 2 \\
        (MaxPool3D) & & \\\hline
            4  & (Nothing set) & & \\
            (Flatten) & & \\ \hline
            5  & Input size & 6912 & 343 \\
            (Linear) & Output size & 128 & 128 \\
                    & Activation & ReLU & ReLU \\\hline
            6  & Input size & 128 & 128 \\
            (Linear) & Output size & 4 & 4 \\
                    & Activation & SoftMax & SoftMax \\\hline

    \end{tabular}
\end{table}

To reduce the size of the final model, we also consider a reduced network, which removes the second convolutional layer and changes the output channels for the first convolutional layer from 32 to one. This reduction in output channels reduces the number of inputs to the first linear layer from 6912 to 343, reducing the number of weights needed to be stored by 95\%.

We also create a model with only a MaxPool3D followed by one linear layer with a SoftMax activation function. This reduces the network to the equivalent of four instances to logistic regression, followed by a SoftMax to compare the outputs.

For training of the network models, we use the Cross Entropy loss and the ADAM optimizer, with $10^{-6}$ as the learning-rate. For the reduced network and Logistic regression, the learning-rate was increased to $10^{-5}$ in order for it to converge to a solution. For each network, four different models were trained using four different seeds for the initiation of the parameters. Each model was trained until the accuracy on the test-set had not increased for 10 consecutive training epochs.

\subsection{Reduced-Precision Neural-Network Parameters}\label{lower-precision-weights}

As discussed before, the available uplink data rates are small, while it is likely that the parameters for the neural network present on the spacecraft will have to be updated after launch. This could be due to lack of original training data, or due to shifts in the input data from aging instruments.

One way to decrease the amount of data to transfer is to store the network weights in a lower precision format. We therefore investigate the effect on the network if the weights are cast to Float16 and BFloat16. We do not put any constraints on the inference, which can still be performed using Float32. Reducing the weights to 16-bit from the original 32-bit should reduce the overall network size by half. To measure the size of the final model, we export the models to ONNX-files and measure the file sizes.

To reduce the network weights even further, we make use of a simple reduction scheme where each parameter is encoded as an 8-bit integer according to:

\begin{equation}
f(x) =
\begin{cases}
    127 & ,\text{if } x\cdot100 > 127 \\
    -127 & ,\text{if } x\cdot100 < -127\\
    \round{x\cdot100} & ,\text{otherwise}
\end{cases}
\end{equation}
Then, before inference, the parameters are then converted back to floating point by dividing by 100. This reduces the range of the parameters to $[1.27,-1.27]$ and the precision to two decimals.

\subsection{Experimental Setup}
Both training and evaluation for the networks were performed on a system with an AMD EPIC 7302P %(Cores: 16, Threads per core: 2)
CPU and an NVIDIA A100 %(SM: 6912)
GPU, running CentOS Linux 8 (Kernel: 4.18.0) with CUDA 12.3. The network, training and testing software was implemented and run using python 3.11.6 with PyTorch 2.2.1. %, GPU Driver version 545.23.08, Nsight Systems 2023.3.3.42, nvidia-smi 545.23.08, nvcc v12.3.52, and gcc v8.5.0.

\section{Results}\label{results}

\subsection{Filtering and Region-of-interest detection} \label{sec:roi_detection}
\begin{figure}[h]
    \centering
    \includegraphics[width=\linewidth]{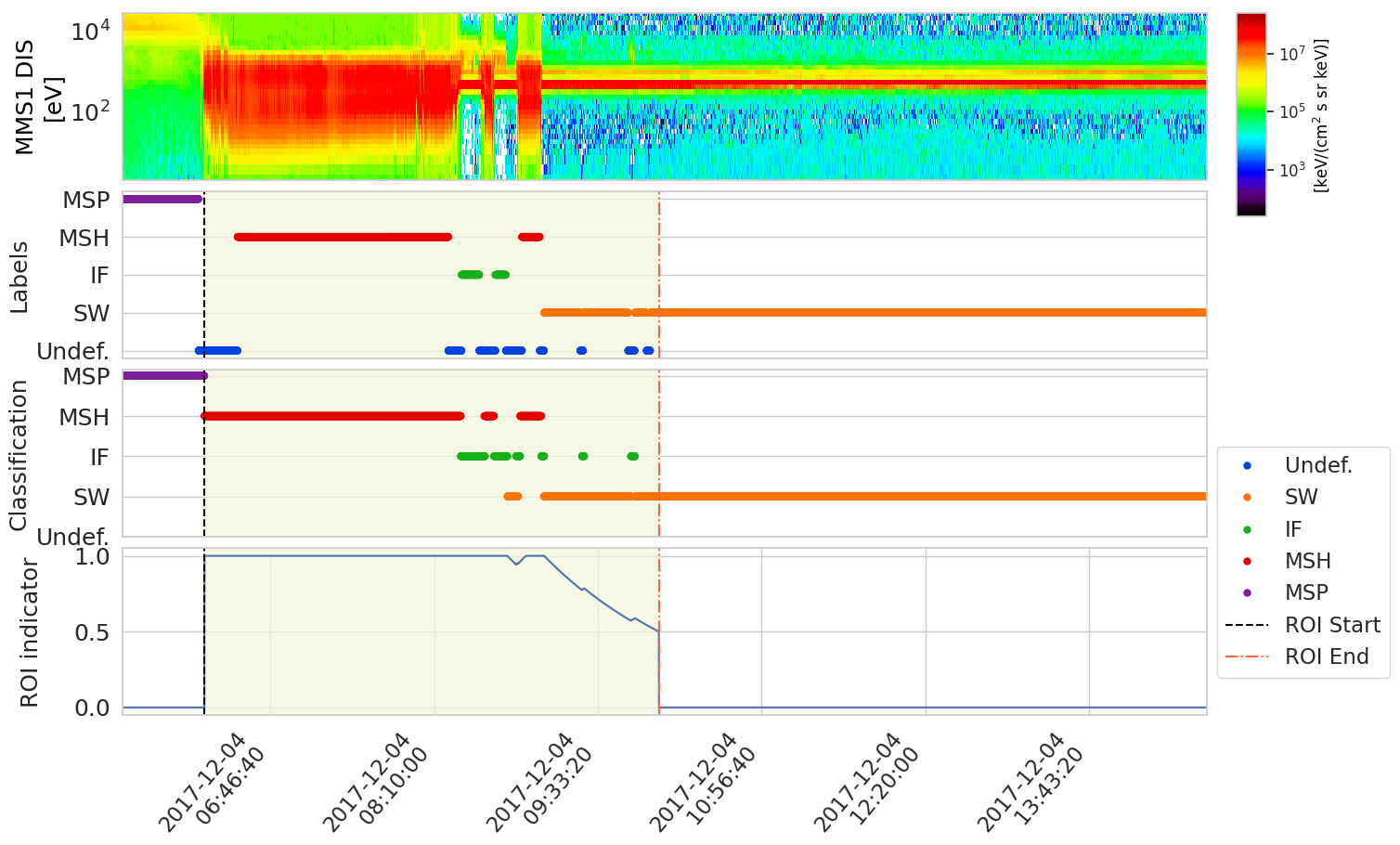}
    \caption{Region of interest detection with baseline Float32 CNN (seed 42). The top most plot is showing the Ion energy spectrum from the FPI data. The second and third plot show the human and CNN classification labels and the bottom plot show the ROI Indicator value from the Algorithm~\ref{alg:roi_classfication}.}
    \label{fig:roi_detection}
\end{figure}

In Fig.~\ref{fig:roi_detection}, we present the results for the region of interest (ROI) detection algorithm, for the configuration in Table~\ref{tab:roi_conf}, when the CNN and ROI detection algorithm is running in our setup on ground. The CNN output and filtering does not categorize any data as Undefined. An example of this can be seen in Fig.~\ref{fig:roi_detection} at the transition from MSP to MSH before 06:46:40, where the region labeled as undefined is either classified as MSP or MSH.
Between 08:10:00 and 09:33:20 we can see how the ROI indicator, the bottom plot in Fig.~\ref{fig:roi_detection}, starts to decrease when the data is classified as SW following a continuous time when it has been classified as either MSH or IF. Then the data is again classified as IF followed by MSH and the ROI indicator increases until it reaches the max value (1.0).
At the end of the marked ROI, the ROI indicator is slowly decaying with small increases when the data is classified as IF. Once the threshold (0.5) is reached, the ROI indicator is set to zero and the region is not marked as part of ROI.

\begin{table}[]
    \centering
    \caption{ROI filter and detection configuration.}
    \label{tab:roi_conf}
    \begin{tabular}{c|c c c c }
        Parameter & $\alpha$ & Threshold & Regions & Decay \\ \hline
        Value & 0.1  & 0.5  & [1,2] & 0.001 \\
    \end{tabular}
\end{table}

Table~\ref{tab:orbit_classifcation} shows how much of each region, based on the labels, falls within the ROI detected by the algorithm for fast survey data from between 2017-12-04 05:31:27 and 2017-12-06 16:22:50. Fig.~\ref{fig:roi_detection} shows a smaller region of this data. We can see that the algorithm performs well at encapsulating the Magnetosheath (MSH) and Ion foreshock (IF), which were the regions specified to the algorithm. It also includes parts of the Solar Wind (SW) and Magnetosphere (MSP) in the ROI, this is mainly due to the decay term extending the ROI when the spacecraft exits the specified regions. As for the Undefined regions, as the classification and filtering does not specify a region as Undefined, whether this is included in ROI depends on how the CNN classifies the data.
\begin{table}[]
    \centering
\caption{The percentage of each human labeled data that falls within the ROI marked by the algorithm for data classified between 2017-12-04 05:31:27 and 2017-12-06 16:22:50 (approximately one orbit.)}
    \label{tab:orbit_classifcation}
    \begin{tabular}{c |c|c c c}
        Network & Label & In ROI & Outside of ROI & Total samples\\  \hline
                & Undefined & 35.21\% & 64.79\% & 4027 \\
        Baseline & SW & 16.07\% & 83.93\% & 5596 \\
         Float32    & IF & 93.84\% & 6.16\% & 406 \\
         (seed 42)  & MSH & 100\% & 0.0\% & 5410 \\
                & MSP & 31.2\% & 68.8\% & 2032 \\ \hline

     %           & Undefined & 35.21\% & 64.76\% & 4027 \\
     %   reduced & SW & 16.05\% & 83.94\% & 5596 \\
     %    int8   & IF & 93.84\% & 6.16\% & 406 \\
     %    (seed 336) & MSH &100\% & 0.0\% & 5410 \\
     %           & MSP & 31.69\% & 68.31\% & 2031 \\ \hline
     %
     %           & Undefined & 35.21\% & 64.76\% & 4027 \\
     %   Logistic& SW & 15.03\% & 84.97\% & 5596 \\
     %    int8 & IF & 93.35\% & 6.65\% & 406 \\
     %    (seed 336) & MSH &100\% & 0.0\% & 5410 \\
     %               & MSP &25.64\% & 74.36\% & 2032 \\ \hline
    \end{tabular}
\end{table}

\subsection{Classification Accuracy}\label{baseline-accuracy}

Table~\ref{tbl:baseline_accuracy} shows a summary of the accuracy on the test set of the four models for each of the three networks. The classification is done according to the output with the highest value and with no further filtering performed. All models were trained to an accuracy higher than 90\% before the accuracy stopped increasing for 10 training epochs. The high accuracy of the smaller Reduced and Logistic models highlights the importance of considering simpler models for any given problem.

\begin{table}[]
    \centering
    \caption{Accuracy on the test set after training for the baseline and reduced Float32 models.} \label{tbl:baseline_accuracy}
    \begin{tabular}{c|c|c|c}
        Network & Seed & Accuracy (\%) & Training Epochs \\ \hline
        \multirow{4}{*}{Baseline} & 42 & 94.15 & 96 \\
                 & 84 & 94.31 & 91 \\
                 & 168 & 94.41 & 118 \\
                 & 336 & 94.91 & 135 \\ \hline
        \multirow{4}{*}{Reduced} & 42 & 94.7 & 145 \\
                 & 84 & 94.7 & 154 \\
                 & 168 & 93.8 & 75 \\
                 & 336 & 94.3 & 82 \\ \hline
        \multirow{4}{*}{Logistic} & 42 & 90.3 & 145 \\
                 & 84 & 95.0 & 167 \\
                 & 168 & 90.2 & 104 \\
                 & 336 & 95.0 & 172 \\
    \end{tabular}
\end{table}

\begin{figure}[h]
    \centering
    \includegraphics[width=0.9\linewidth]{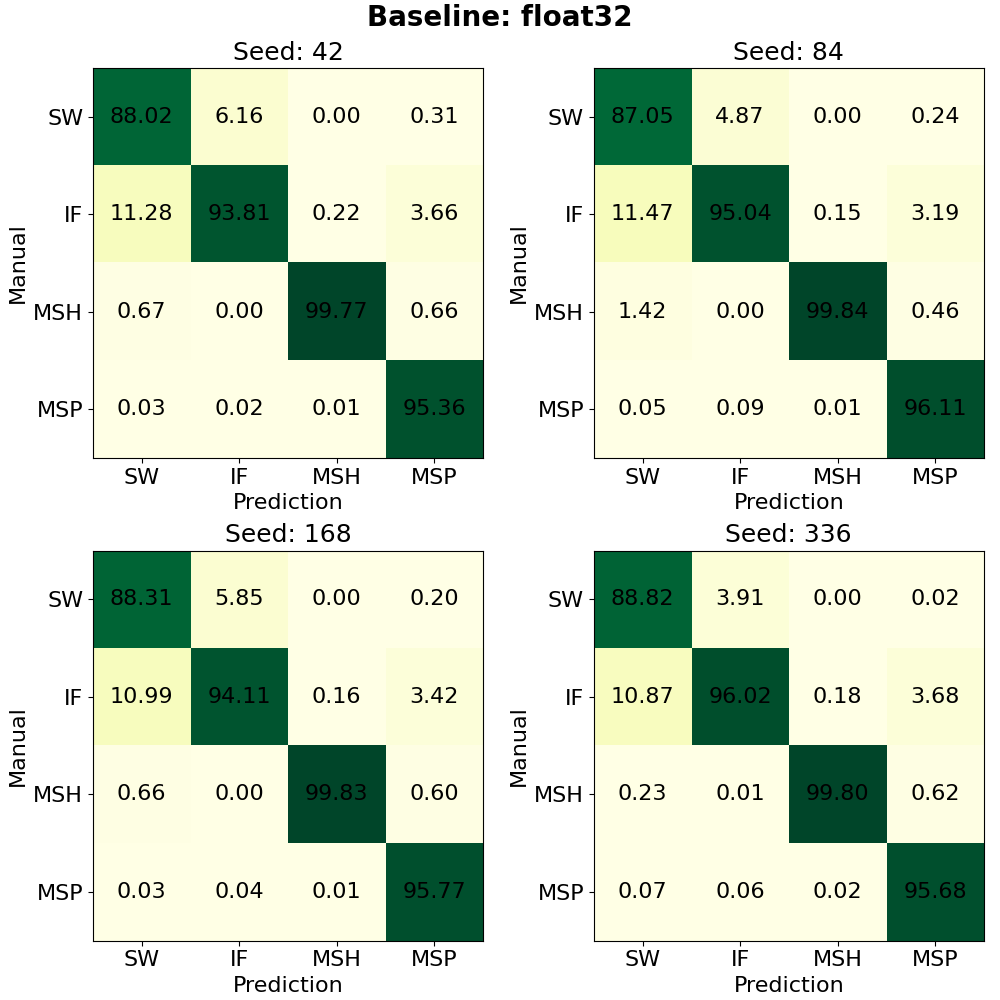}
    \caption{Confusion matrices for the baseline Float32 network.}
    \label{fig:baseline_confusion_matrix}
\end{figure}

Investigating the confusion matrices for the baseline Float32 models in Fig.~\ref{fig:baseline_confusion_matrix} we can see that the models have a hard time separating the solar wind from the ion foreshock. For more than 10\% of the samples, the model is classifying data labeled as Solar Wind as Ion Foreshock.

\begin{figure}[h]
    \centering
    \includegraphics[width=0.9\linewidth]{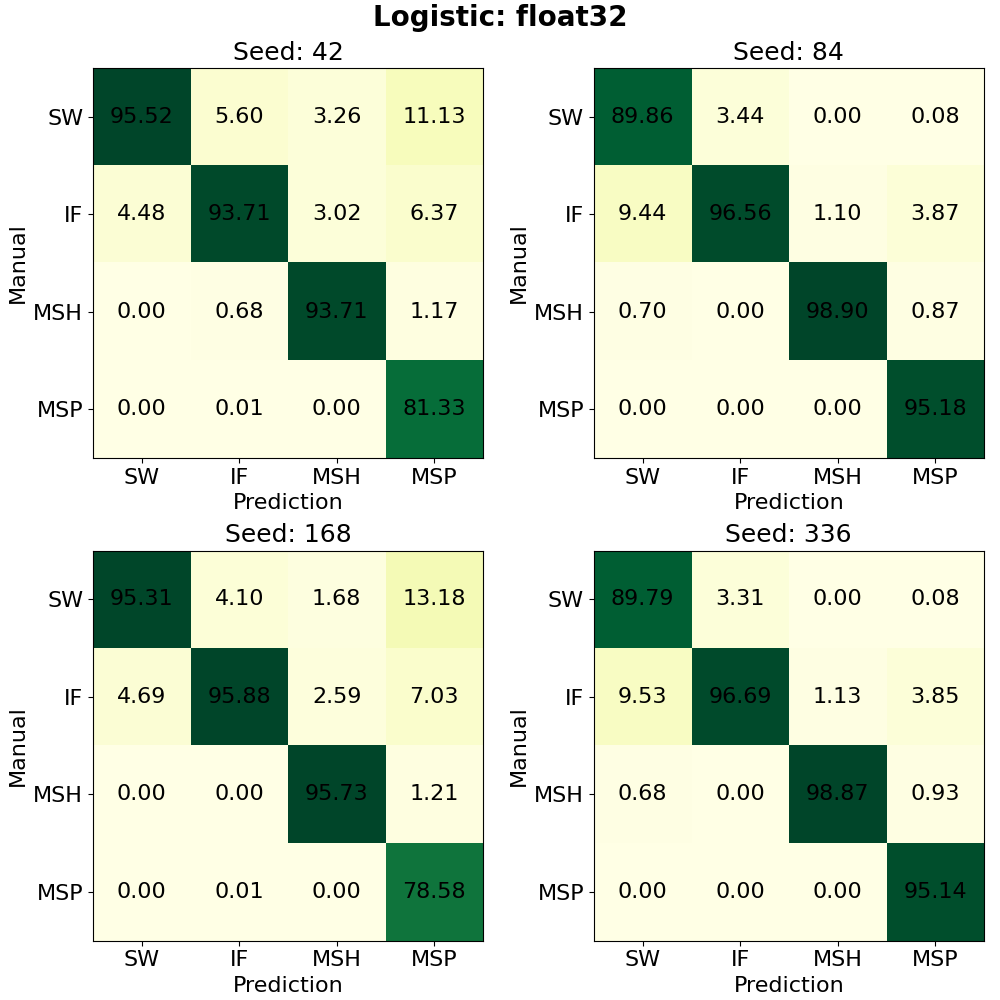}
    \caption{Confusion matrices for the logistic Float32 network.}
    \label{fig:logistic_confusion_matrix}
\end{figure}

The Logistic models appear to have fallen into two different local minima: one with 90\% accuracy and another one with 95\%. Looking at the confusion matrices for these models in Fig.~\ref{fig:logistic_confusion_matrix}, we find results very similar to the Baseline models in two of the seeds, 84 and 336. However, the seeds 42 and 168 have a lower overall accuracy and tend to mislabel solar wind or ion foreshock as magnetosphere. Where more than 17\% of the data classified as MSP by the model is actually labeled as SW or IF.

\subsection{Reduced-Precision Neural-Network Parameters}\label{sec:lower-precision-parameters}

\begin{figure}[h]
    \centering
    \includegraphics[width=\linewidth]{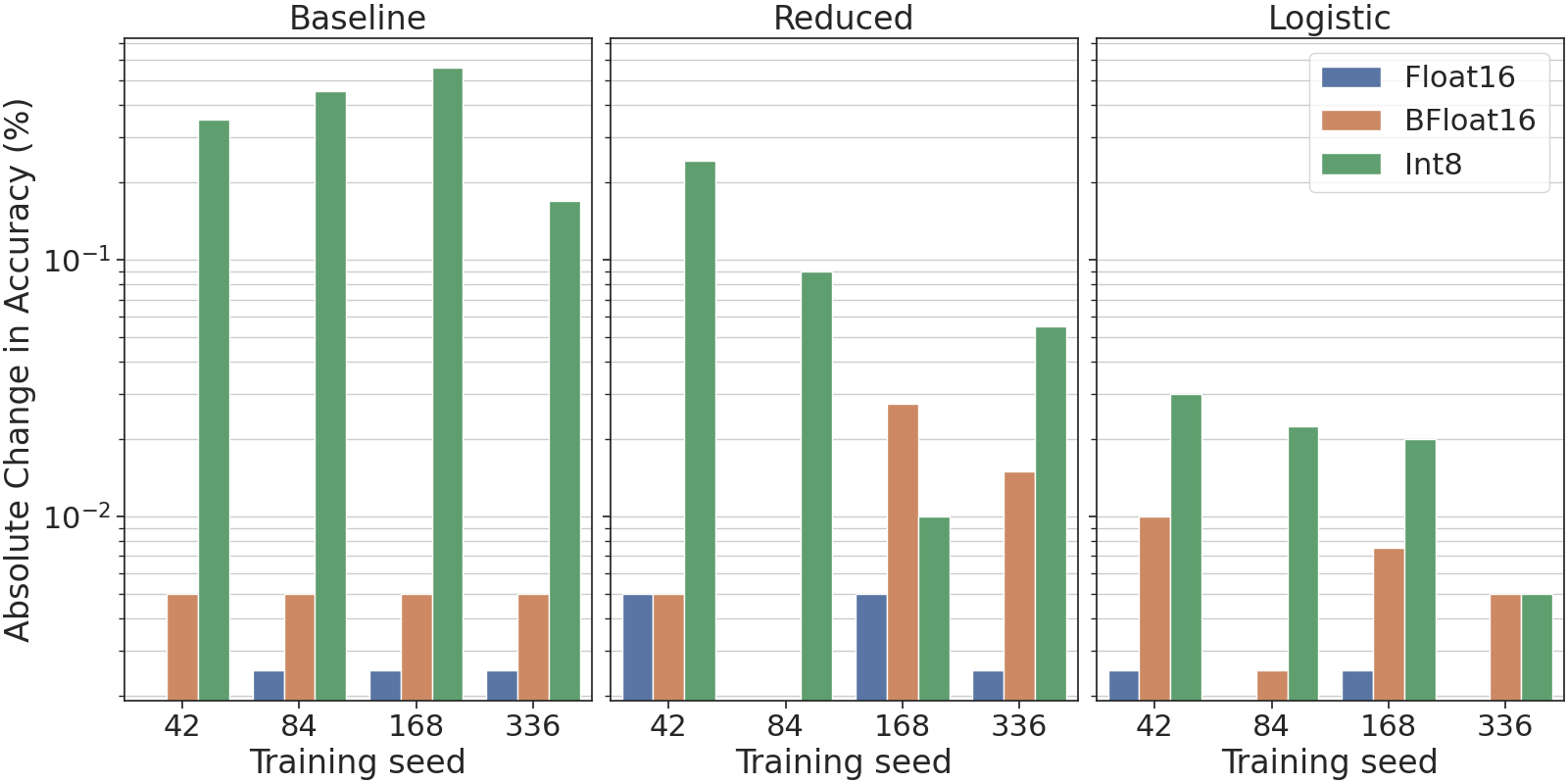}
    \caption{The absolute change in accuracy between the reduced precision parameters and full precision Float32 parameters.}\label{fig:acc_diff}
\end{figure}

Converting the data to lower precision implies that we lose some information due to round-off error. It is therefore important to evaluate how this affects the final classification of the network. As can be seen in Fig.~\ref{fig:acc_diff} the change in accuracy is less than 0.6 percentage point for all formats. What is interesting to note is that the smaller the network, the smaller the effects of the Int8 format on the accuracy. The y-axis of Fig.~\ref{fig:acc_diff} is scaled logarithmically to make the accuracy changes for Float16 visible.

There is a loose coupling between the round-off error and the change in accuracy. This can be seen when Fig.~\ref{fig:acc_diff} with the Round-off error in Table~\ref{tab:round-off_error}. We can see that for the Baseline Int8 model, we have the largest change in accuracy and the largest worst case round-off error. However, as the classification is defined by the highest output from the network, we can have changes in output that does not affect the final classification. This can be seen in that when comparing the different seeds for the same data type, the seed with the largest round-off error does not always have the largest change in accuracy.

\begin{table}[]
    \centering
    \caption{Euclidean norm of the round-off error for the different number formats. Only the worst models for each network and format are presented.}
    \label{tab:round-off_error}
    \begin{tabular}{c|c|c|c}
        Network & Data type & Seeds & Round-off error \\
                &           &       & Euclidean norm \\ \hline
        \multirow{3}{*}{Baseline} & Float16 & 42,84,168,336 & 0.0019 \\
                 & BFloat16 & 336 & 0.0156 \\
                 & Int8 & 336 & 2.7245 \\\hline
        \multirow{3}{*}{Reduced} & Float16 & 84 & 0.0027 \\
                 & BFloat16 & 84 & 0.0218 \\
                 & Int8 & 168 & 0.6128 \\\hline
        \multirow{3}{*}{Logistic} & Float16 & 336 & 0.0024 \\
                 & BFloat16 & 42, 336 & 0.0181 \\
                 & Int8 & 336 & 0.2633 \\
    \end{tabular}
\end{table}

\subsection{Neural Network Sizes}
In Table~\ref{tab:round-off_error}, we have the file sizes when the network is exported to ONNX files. As PyTorch does not support converting models to Int8\footnote{Not counting the quantization framework, \url{https://pytorch.org/docs/stable/quantization.html}}, the Int8 file sizes are calculated based on the Float32 and Float16 sizes. In the table, we can see that the change in format for the parameters can have a large impact on the file size and thereby the upload times, especially for larger networks. As expected, reducing number representation from Float32 to Float16 decreases file sizes by half for the Baseline network. Reducing the parameters even further to an 8-bit representation would reduce the network to approximately a fourth of the original size.

The size reductions for the reduced and logistic network, compared to their Float32 variants, are not as large. This is due to the lower number of network parameters, which thereby constitute a smaller part of the overall size of the file. However, the network parameters are still the major factor for the file sizes.

\begin{table}[]
    \centering
    \caption{Size of ONNX file. ($^*$Int8 calculated based on Float32 and Float16 sizes.)}\label{tbl:parameter_sizes}
    \begin{tabular}{c|c|c c c}
                & & & & Approx.\\
       Network & Data format & Size (MB) & Relative & Upload time \\
               &             &           &  & (2kbit/s)\\ \hline
        \multirow{4}{*}{Baseline} & Float32 & 3.668 & 1 & 4h\\
                 & Float16 & 1.847 & 0.5 & 2h\\
                 & BFloat16 & 1.847 & 0.5 & 2h\\
                 & Int8 & 0.937$^*$ & 0.26 & 1h\\ \hline
        \multirow{4}{*}{Reduced} &  Float32 & 0.193 & 0.05 & 13 min\\
                 & Float16 & 0.103 & 0.03 & 7 min\\
                 & BFloat16 & 0.103 & 0.03 & 7 min\\
                 & Int8 & 0.058$^*$ & 0.016 &  4 min\\ \hline
        \multirow{4}{*}{Logistic} &  Float32 & 0.042 & 0.011 & 3 min\\
                 & Float16 & 0.025 & 0.007 & 2 min\\
                 & BFloat16 & 0.025 & 0.007 & 2 min\\
                 & Int8 & 0.017$^*$ & 0.005 &  1 min\\
    \end{tabular}
\end{table}

\section{Related Work}\label{related-work}

\textcite{caballeroInferenceRecyclableObjects2021} and \textcite{stackerDeploymentDeepNeural2021} show that neural networks can be quantized to Float16 and Int8 for deployment of object detection networks on edge devices. They show that by quantizing the weights, they decrease the inference time without significant loss in accuracy.

\textcite{guptaDeepLearningLimited2015} show that neural networks can be trained to the same performance as 32-bit Float, using a 16-bit fixed-point format if stochastic rounding is used. \textcite{kalamkarStudyBFLOAT16Deep2019} show that BFloat16 can be used to train different types of neural networks, CNN among those, to the same accuracy as Float32.

\textcite{giuffridaFSat1MissionFirst2022} showcased the use of neural network onboard $\phi$-Sat 1 for cloud detection in earth observation data.

Similarly, \textcite{mateo-garciaInorbitDemonstrationRetrainable2023} demonstrated the use of neural network in space for detection of flooding. They showed how a model could be updated in orbit after launch to improve the performance. %One important factor in the model selection was the size, due to the small uplink capacity of the spacecraft. %SM: removed for space

\section{Discussion and Conclusions}

The classification filtering and algorithm for ROI detection presented here is basic and therefore robust for space applications. As can be seen in Sec.~\ref{sec:roi_detection} it works well to detect the specified region of interest. In addition to this, its responsiveness can be updated by changing three parameters and it can be set to detect any combination of the regions by changing a fourth parameter. This number of parameters is significantly lower than the thousands needed to update the entire network models. In addition to this, the algorithm is independent of what the data classification represents and can be used for any time series of classified data points. However, if the area of interest is not in one of the regions, but at the edge as in the case of magnetic reconnection in the magnetopause, there is a risk that it will miss it when entering the ROI. The decay of the indicator value will, on the other hand, ensure that once a region has been detected the ROI will remain even if there are some spurious classifications not specified as interesting. It will thereby capture the edge region at the end of a ROI.

The classification approach in this work is reactive, classifying data when it has been collected and using the classification in an algorithm. However, for future work it would be interesting to investigate a more proactive approach where, for example, transformers could be used to predict events where collection of data at a higher-rate is desirable.

The MMS FPI data used for training and classification is intended for scientific investigations. It is therefore highly post processed after downloaded to ground and is not fully representative of what would be available on the spacecraft. It is possible that with less post-processed data containing more artifacts, the simpler models presented here will not be as robust. To continue this work, we would need to consider rawer, less post-processed, data from the instruments.

We did not consider any compression of the final ONNX-file, this is one way that could be used to further reduce the file sizes. However, the reduction of the parameters to Float16 and Int8 representations can be considered lossy compressions. By reducing the number of bits used to represent the parameters, we can reduce the size of the file to transfer without needing any decompression step on the spacecraft. Furthermore, by using a smaller network, we reduce the overall number of parameters and the file size even further. As we have shown here, this does not need to come at any loss of accuracy. This highlights the importance of considering smaller networks for solving problems where resources are limited and not just using techniques to reduce the model sizes of large networks.

A smaller network also has the added benefit of requiring fewer calculations, which can speed up the inference and reduce power consumption. This is a very desirable trait, as the onboard power is a limited resource, especially further out from the sun. However, quantifying the power usage of a given model will be dependent on the hardware and has to be investigated for different alternatives of MA/AI Processing Units together with network topologies and quantization levels.

For compression of the parameters to an 8-bit format, we used a simple conversion method in this work. This indicated that the parameters can be compressed to 8-bits for transfer to the spacecraft, without significant loss of performance during inference. Although, the actual inference step was still performed in Float32. There are, however, more sophisticated quantization schemes where also the inference steps can be performed using the quantized parameters. These have been shown to increase the inference performance on edge devices. This is an interesting option to evaluate for use onboard spacecraft. It should also be noted that the reduction to the 8-bit format introduces sparsity into the network parameters as some parameters are rounded to zero, this could lend itself to further compression by using for example Compressed Sparse Row (CSR), or similar formats.

The change in accuracy observed for the different models when the precision of the parameters are reduced is very low, less than 0.6 percentage points, for all the models. However, as can be seen in Fig.~\ref{fig:acc_diff}, the standard deviation compared to the mean is large. To fully understand the effect of low precision on the networks, a more extensive uncertainty analysis will have to be performed. This could be done by training more models to a given accuracy, either with different seeds, as has been done here, or using different training-sets. This could then be used to quantify the mean change in accuracy and the standard deviation.

In this work, we have shown how you can detect a region of interest in a series of data classifications using a simple algorithm operating on a time series of data classifications. Detecting over 99\% of the regions specified as interesting, as part of the region-of-interest. It also allows for tuning, making it more or less reactive, using only three parameters and allows for changing the regions of interest by using a fourth parameter.

We have also shown how an existing model for classifying Ion distribution data can be simplified to a reduced model with only one convolutional layer or a logistic model, with only a single linear layer. This implies reducing the size by 95.0\% and 98.9\%, respectively, compared to the original model, without reducing the accuracy of the final predictions.

We have further shown that the sizes of all the models can be reduced by quantizing the network parameters to lower precision formats. By reducing the parameters to Float16 and Int8 formats, the models can be reduced by up to 50\% and 75\% percent compared to their Float32 variant. The reduction in precision for the storage of the parameters changes the accuracy of the network by less than 0.6 percentage points.

\printbibliography

\end{document}